\documentclass[10pt,twocolumn,letterpaper]{article}

\usepackage{iccv}              % To produce the CAMERA-READY version
\usepackage{times}
\usepackage{epsfig}
\usepackage{graphicx}
\usepackage{amsmath}
\usepackage{amssymb}
\usepackage[accsupp]{axessibility}  % Improves PDF readability for those with disabilities.
\usepackage{booktabs}
\usepackage{multicol}
\usepackage{multirow}
\usepackage{makecell}
\usepackage{colortbl} 
\usepackage{xcolor}
\usepackage{array}
\usepackage{arydshln}
\usepackage[accsupp]{axessibility}

\definecolor{iccvblue}{rgb}{0.21,0.49,0.74}
\usepackage[pagebackref,breaklinks,colorlinks,allcolors=iccvblue]{hyperref}

\def\paperID{11597} % *** Enter the Paper ID here
\def\confName{ICCV}
\def\confYear{2025}

\title{Q-Frame: Query-aware Frame Selection and Multi-Resolution Adaptation for Video-LLMs}

\author{Shaojie~Zhang$^{1*}$~~~Jiahui~Yang$^{1*}$~~~Jianqin~Yin$^{1\dagger}$~~~Zhenbo~Luo$^{1\ddagger}$~~~Jian~Luan$^{1}$\\
    {\normalsize $^{1}$MiLM Plus, Xiaomi Inc.} \\
    {\tt\small zhangshaojie5@xiaomi.com, yangjiahui1@xiaomi.com, jianqinyin4@gmail.com} \\
    {\tt\small luozhenbo@xiaomi.com, luanjian@xiaomi.com}
}
\begin{document}
\twocolumn[{%
\renewcommand\twocolumn[1][]{#1}%

\maketitle%

\vspace{-0.5in}%
\begin{center}
\includegraphics[width=\linewidth]{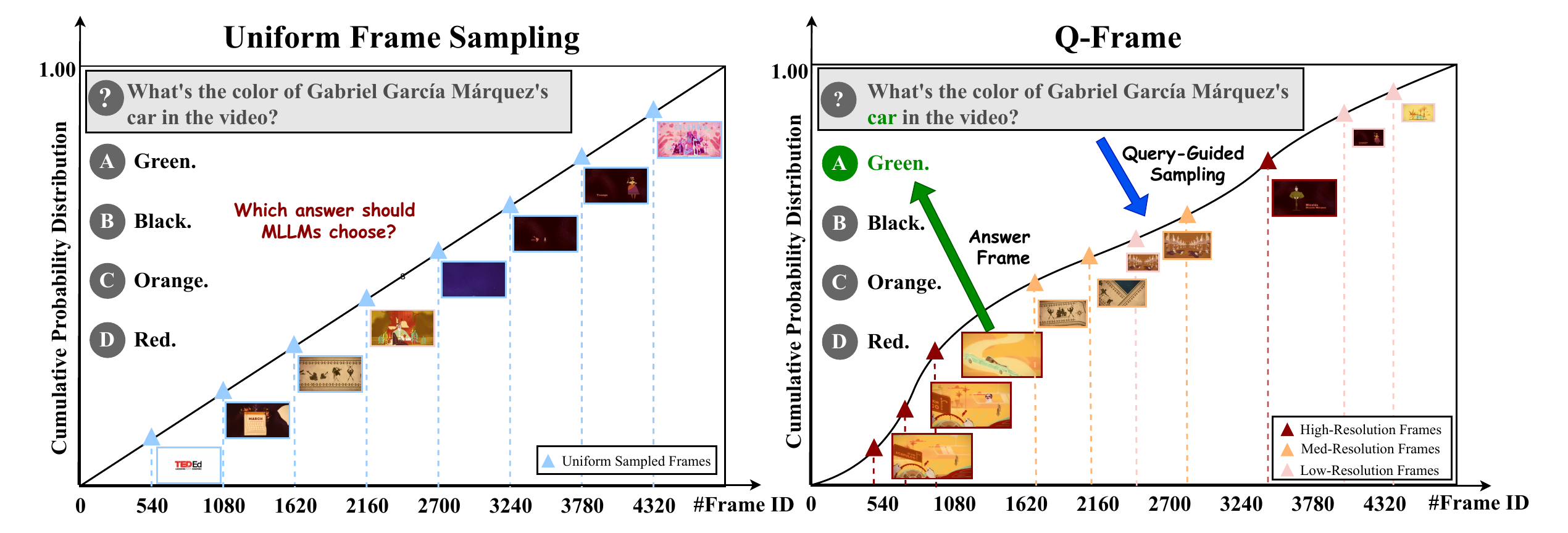}
        \vspace{-0.3in}
   \captionof{figure}{Comparison of uniform frame sampling and proposed \textbf{Q-Frame} sampling. Uniform Frame Sampling selects frames at fixed intervals, leading to sparse and potentially irrelevant frame selections that disrupt temporal continuity. In contrast, Q-Frame dynamically selects query-aware frames and adapts their resolution, ensuring that the most relevant frames are selected with optimized resolution to preserve crucial visual details. This adaptive approach enhances the Video-LLMs' ability to understand long-form videos more efficiently and effectively, addressing the limitations of traditional frame sampling methods.
   } 
\label{fig:motivation}
\end{center}
}]

\maketitle

{\let\thefootnote \relax \footnote{$\dagger$ Corresponding author; $\ddagger$ Project leader; $^*$ Equal contribution.}}

\begin{abstract}
    Multimodal Large Language Models (MLLMs) have demonstrated significant success in visual understanding tasks. However, challenges persist in adapting these models for video comprehension due to the large volume of data and temporal complexity. Existing Video-LLMs using uniform frame sampling often struggle to capture the query-related crucial spatiotemporal clues of videos effectively. In this paper, we introduce Q-Frame, a novel approach for adaptive frame selection and multi-resolution scaling tailored to the video's content and the specific query. Q-Frame employs a training-free, plug-and-play strategy based on a text-image matching network like CLIP, utilizing the Gumbel-Max trick for efficient frame selection. Q-Frame allows Video-LLMs to process more frames without exceeding computational limits, thereby preserving critical temporal and spatial information. We demonstrate Q-Frame's effectiveness through extensive experiments on benchmark datasets, including MLVU, LongVideoBench, and Video-MME, illustrating its superiority over existing methods and its applicability across various video understanding tasks.
\end{abstract}    
\section{Introduction}
\label{sec:intro}

Multimodal Large Language Models (MLLMs) have achieved remarkable performance in a wide range of visual understanding tasks, showcasing their potential across various domains \cite{bai2023qwen, openai2024hello, wang2024qwen2}. Building upon these advancements, MLLMs are now poised to tackle more complex challenges, such as video understanding, which requires not only capturing spatial details but also effectively modeling temporal dynamics \cite{cheng2024videollama, li2024mvbench, lin2024video, maaz2023video}. 

Video understanding, however, presents unique challenges due to the inherently large number of video frames and the continuous nature of video content. Despite progress in video understanding, current methods suffer from inherent limitations due to the temporal redundancy in videos and the restricted context length within MLLMs. A typical video contains a large number of frames—e.g., a 3-minute video at 24 frames per second (fps) results in approximately 4,320 frames. Modern MLLMs remain constrained by their context—VideoLLaMA2 \cite{cheng2024videollama} supports only 2,000 tokens, while VILA-V1.5 \cite{lin2024vila} extends this to roughly 4,000 tokens. It becomes clear that using all frames for understanding is computationally infeasible. Previous approaches have relied on uniform frame sampling, where a fixed subset of frames is selected at regular intervals. While this method is straightforward, it has several drawbacks: it results in \textbf{sparse frame selections} that break temporal continuity, it follows a \textbf{static and query-agnostic} sampling strategy, and it \textbf{treats all frames equally}, which is inefficient and suboptimal for fine-grained tasks \cite{lin2024video, maaz2023video, zhang2024video, wang2024internvideo2}.

Specifically, the inherent sparsity in uniform sampling is particularly problematic for long videos. As video length increases, the fixed number of sampled frames becomes increasingly sparse, disrupting the temporal flow and potentially omitting critical transitions and subtle details. This is detrimental for tasks such as object counting and temporal grounding \cite{fu2024video}. Furthermore, uniform frame sampling is inherently static, meaning that the same combination of frames is adopted for each question, regardless of the query's particular needs. Some efforts, such as text-frame matching \cite{liang2024keyvideollm} and frame ranking \cite{yu2024frame}, have been introduced to break this static selection. However, these methods fundamentally fail to capture the intricate temporal dependencies and dynamic contextual relationships inherent in advanced video comprehension tasks. In the end, the distribution of information density in video is often not uniform, and uniform frame sampling fails to consider the varying information density across frames. Downsampling high-resolution frames can lead to a loss of fine-grained details, which negatively impacts the Video-LLMs' ability to capture spatial information. While maintaining the native resolution of each frame will significantly increase the consumption of computing resources. This one-size-fits-all approach may fail to capture spatiotemporal clues in videos.

\begin{figure}[htbp]
    \centering
    \includegraphics[width=0.95\linewidth]{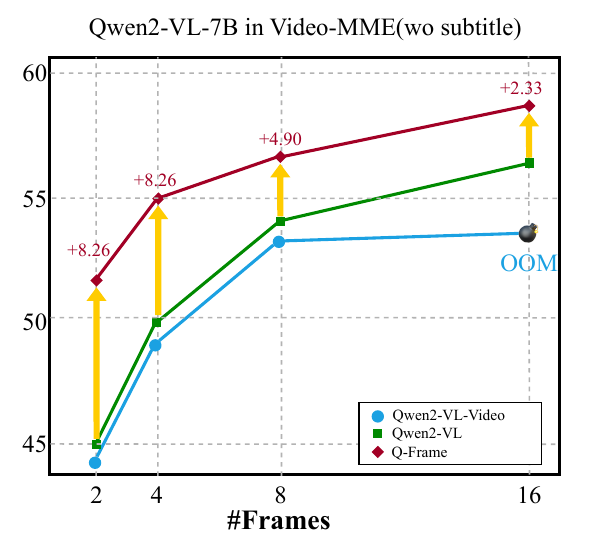}
    \vspace{-10pt}
    \caption{Overall accuracy (\%) Video-MME (without subtitles) for Qwen2-VL, comparing uniform sampling and Q-Frame selected from 128 frames. Note that Qwen2-VL-Video is the video understanding model, and Qwen2-VL is a multi-image understanding model based on different activation weights.}
    \label{fig:comparsion}
\end{figure}

To address these limitations, we propose \textbf{Q-Frame}, a novel framework for query-adaptive frame selection and resolution adaptation in Video-LLMs. As illustrated in Figure~\ref{fig:motivation}, Q-Frame dynamically selects the most relevant frames from a video and adjusts their resolutions according to both the video content and the given query. In contrast to prior approaches that treat all frames uniformly, Q-Frame introduces a dynamic strategy that makes frame sampling \emph{query-aware} and computationally efficient. This strategy is derived from a discrete probability distribution parameterized by the output of a text–image matching model (e.g., CLIP \cite{radford2021learning}). By leveraging the Gumbel-Max trick \cite{jang2016categorical}, we can sample from this distribution to obtain an optimal combination of frames without resorting to complex optimization or any additional training, making Q-Frame easy to implement. Importantly, as shown in Figure \ref{fig:comparsion}, Q-Frame allows a Video-LLM to process a greater number of informative frames without exceeding the computational budget, thus preserving crucial visual details that would be lost with uniform sampling. Moreover, Q-Frame is model-agnostic and plug-and-play, requiring no fine-tuning of the Video-LLM and can be applied to both open-source models \cite{wang2024qwen2, li2024llava} and closed-source API-based models \cite{openai2024hello}.

The main contributions of this paper are as follows:
\begin{enumerate}
    \item We introduce \textbf{Q-Frame}, a novel framework for query-aware frame selection and multi-resolution frame scaling that improves long-form video understanding by focusing on the most relevant visual content for a given query.
    \item We propose a training-free, plug-and-play mechanism that leverages a CLIP-based vision-language model to guide frame selection and resolution assignment \emph{without} any additional model training or fine-tuning.
    \item We validate the effectiveness of Q-Frame through extensive experiments on three benchmark datasets (MLVU \cite{zhou2024mlvu}, LongVideoBench \cite{wu2025longvideobench}, and Video-MME \cite{fu2024video}), demonstrating its superiority over previous methods and its practical usability for video understanding tasks.
\end{enumerate}

%-------------------------------------------------------------------------
\section{Related Work}

\textbf{Video Large Language Models.} 
Early video understanding approaches primarily relied on handcrafted features or conventional deep networks with limited capacity to capture temporal dynamics and semantic richness in visual content \cite{tang2023video}. Recent advances in Multimodal Large Language Models (MLLMs) have revolutionized video understanding. The Video-LLMs can be categorized into the following three categories. Firstly, VideoChat \cite{zhang2023video}, LLaVA-Hound \cite{zhang2024direct}, InternVideo2 \cite{wang2024internvideo2}, ShareGPT4Video \cite{chen2025sharegpt4video}, Video-LLaVA \cite{lin2024video}, and LLaVA-Video \cite{zhang2024video} leverage synthesized video-instruction data (e.g., video captions and open-ended QA) to enhance fine-grained action understanding and narrative tracking. Secondly, Video-ChatGPT \cite{maaz2023video}, MovieChat \cite{song2024moviechat}, TimeChat \cite{ren2024timechat}, Video-LLaMA \cite{zhang2023video}, and Video-CCAM \cite{fei2024video} develop efficient feature compression methods for holistic video semantics, employing temporal aggregation and contrastive spatial-temporal modeling. Finally, LongVA \cite{zhang2024long}, LongVILA \cite{xue2024longvila}, and Kangaroo \cite{liu2024kangaroo} extend the LLM's context length to accommodate extended visual token sequences for long video analysis.

\textbf{Keyframe Selection.} 
Keyframe selection is a fundamental technique for identifying representative frame combinations that balance content coverage and reduce redundancy. In the context of early action recognition, AdaFrame \cite{wu2019adaframe} pioneered reinforcement learning (RL)-based adaptive frame selection, later extended by \cite{wu2019multi} through multi-agent decision-making. Dynamic video analysis has led to the development of motion-guided approaches such as MGSampler \cite{zhi2021mgsampler} and its probabilistic variant PMGSampler \cite{bai2025robust}, which prioritize motion-salient frames. Most Video-LLMs \cite{lin2024video,maaz2023video,zhang2024video,wang2024internvideo2} adopt query-agnostic uniform sampling, often resulting in incomplete relevant content capture. Recent advances introduce query-conditioned strategies. KeyVideoLLM \cite{liang2024keyvideollm} implements query-aware top-K semantic retrieval, whereas Frame-Voyager \cite{yu2024frame} employs loss-driven combinatorial optimization. Nevertheless, existing methods face persistent challenges in temporal fragmentation and rigid resolution constraints during frame representation.

\textbf{Any-Resolution Processing.}
MLLMs typically encode visual inputs as discrete visual tokens for cross-modal alignment. Seminal works \cite{liu2023visual,liu2024improved} establish this paradigm through fixed-resolution vision encoders, inherently constraining input flexibility. Subsequent approaches \cite{li2024monkey,chen2024internvl,liu2024llavanext} address native resolution preservation by partitioning images into fixed-grid patches, yet retain structural rigidity and disregard spatial dependencies during tokenization. A breakthrough emerges with Qwen2-VL's \cite{wang2024qwen2} Native Dynamic Resolution framework, enabling adaptive visual token generation across varying input resolutions. Although any-resolution techniques have been well-researched for Image-LLMs, the dynamic resolution adaptation of Video-LLMs remains unexplored.
%-------------------------------------------------------------------------
\section{Method}
\subsection{Overview}
This work focuses on Video Question Answering (VQA) as the primary task while ensuring generalization to various video understanding scenarios, such as summarization and temporal grounding \cite{yu2024frame}. The Video-LLM pipeline can be formalized as ($\mathcal{V}$, \textit{q}) $\Rightarrow$ \textit{a}, where the input video $\mathcal{V} = \{ \text{Frame}^i\}_{i=1}^D$ with $D$ frames, and \textit{q} represents the textual query. 

\begin{figure*}[htbp]
    \centering
    \includegraphics[width=\linewidth]{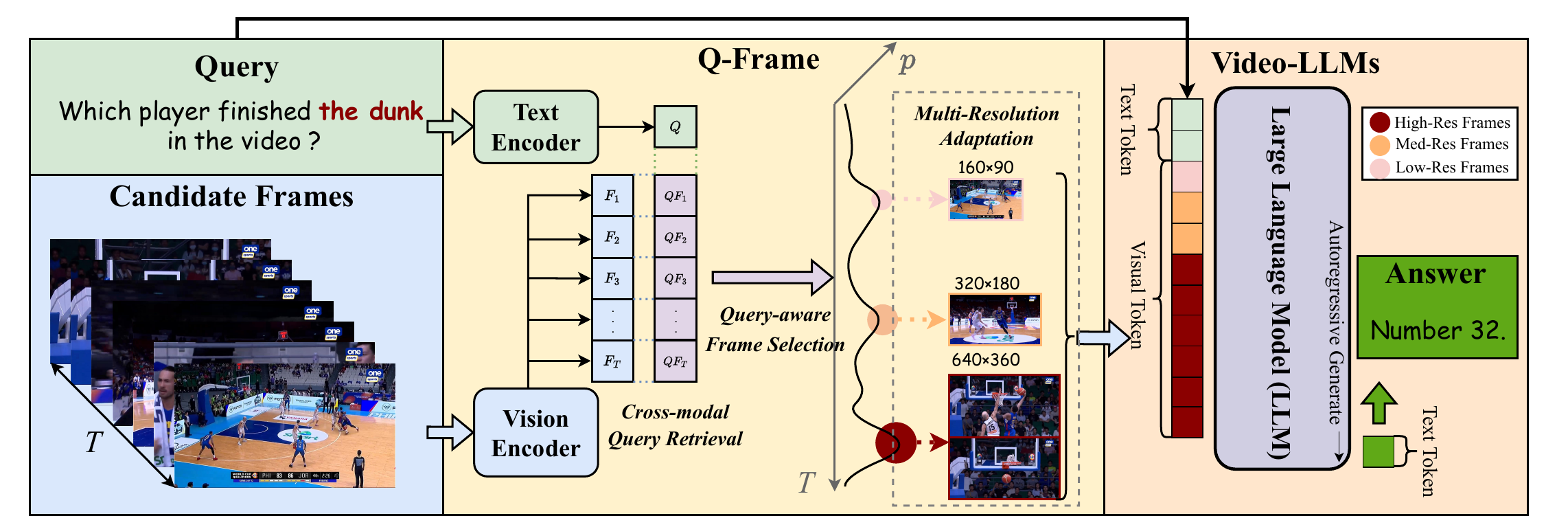}
    \vspace{-10pt}
    \caption{The overall framework of Q-Frame. Q-Frame is composed of Cross-modal Query Retrieval (CQR), Query-Aware Frame Selection (QFS), and Multi-Resolution Adaptation (MRA). CQR focuses on retrieving the most semantically relevant frames from the video based on the textual query, ensuring that only meaningful visual information is considered. QFS is designed to adaptively select frames based on their relevance to the query, enhancing efficiency by concentrating on the most important temporal segments. MRA aims to optimize computational resources by assigning varying resolutions to frames, preserving fine details in important frames while reducing costs for less critical ones. It should be noted that the preprocessing strategy of Video-LLMs is different, and the MRA in the dotted line is not applicable to every model.}
    \label{fig:framework}
\end{figure*}

The proposed Q-Frame framework, as illustrated in Figure \ref{fig:framework}, adapts dynamically to select query-aware frame sequences with resolution scaling. It enables Video-LLMs to process extended frame sequences while maintaining a fixed computational budget through two key mechanisms: 1) content-aware frame selection based on cross-modal query relevance and 2) resolution scaling that optimally allocates visual tokens across spatial dimensions. Specifically, Q-Frame is composed of Cross-modal Query Retrieval (CQR), Query-Aware Frame Selection (QFS), and Multi-Resolution Adaptation (MRA). CQR focuses on retrieving the most semantically relevant frames from the video based on the textual query, ensuring that only meaningful visual information is considered. QFS is designed to adaptively select frames based on their relevance to the query, enhancing efficiency by concentrating on the most important temporal segments. MRA aims to optimize computational resources by assigning varying resolutions to frames, preserving fine details in important frames while reducing costs for less critical ones.

\subsection{Cross-modal Query Retrieval}
\label{sec:3_2}
To effectively answer fine-grained associations of objects, actions, and scenes, it is necessary to retrieve key frames that semantically match the query. We first downsample the original video $\mathcal{V}$ to obtain a candidate frame sequence $\mathcal{F} = \{\text{Frame}^j \}_{j=1}^T \subset \mathcal{V}$ using uniform sampling, where $T \ll D$ specifies the target frame budget.

We leverage pre-trained CLIP-like models that align millions of image-text pairs, offering excellent zero-shot capabilities for cross-modal retrieval \cite{radford2021learning}. The pre-trained Vision-Language Model (VLM) maps the visual input $\mathcal{F}$ and textual input $q$ into a shared semantic space:
\begin{equation}
\begin{split}
    & Q = \text{VLM}_{\text{text}}(q) \in \mathbb{R}^d, \\
    & F = \text{VLM}_{\text{vision}}(\mathcal{F}) \in \mathbb{R}^{T \times d}
\end{split}
\end{equation}
where $\text{VLM}_{\text{vision}}(\cdot)$ and $\text{VLM}_{\text{text}}(\cdot)$ denote the vision and text encoders of the VLM, respectively. To compute the semantic relevance of each frame, we compute the similarity as the inner product between the query and frame embeddings:
\begin{equation}
    I = QF^{\text{T}} \in \mathbb{R}^{1 \times T}
\end{equation}

\subsection{Query-aware Frame Selection}
\label{sec:3_3}
CLIP-like models are primarily trained on image-text pairs, which inherently neglect the temporal correlations present in video frames. To address this, we introduce a probability-guided sampling strategy based on the Gumbel-Max trick \cite{jang2016categorical}, which efficiently samples frames according to their query-relevant matching intensities. This strategy not only ensures diversified frame selection but also dynamically adjusts the probability ordering, enhancing the robustness of the sampling strategy.

First, we convert the matching intensity $I$ into a probability distribution using temperature scaling:
\begin{equation}
    \pi = \text{Softmax}(I/\tau)
\end{equation}
which indicates the probability that each frame is selected, where $\tau$ is the temperature parameter controlling the sharpness of the distribution. 

Subsequently, independent Gumbel noise is injected into the logarithm of the probability distribution to perturb the log-probabilities:
\begin{equation}
\begin{split}
    g &= - \text{log}\ (- \text{log} \ \epsilon), \epsilon \in U[0,\ 1]^{T} \\
    p &= \log \pi + g
\end{split}
\end{equation}
where $g$ follows a standard Gumbel distribution. Using the Gumbel-Max trick, we draw discrete samples from the categorical distribution:
\begin{equation}
    \text{idx}^{\text{select}} = \{ i \ | \ \text{rank}(i) \leq K \}
\end{equation}
Here, $\text{rank}(\cdot)$ sorts the frames by the value of $p$, and $\text{idx}^{\text{select}}$ retrieves the indices of the top-K frames after noise injection. This sampling method combines the exploration-exploitation trade-off, allowing for a more flexible and robust frame selection.

\subsection{Multi-Resolution Adaptation}
Many existing Video-LLMs process all frames at the same resolution \cite{maaz2023video,lin2024video,zhang2024video}. However, frames that contain fine details (e.g., small objects or textual overlays) benefit from higher resolution, while less critical frames can be processed at lower resolutions to reduce computational cost. To address this, we propose a multi-resolution strategy that adapts frame resolutions based on their relevance to the query.

Using the query-aware sampling strategy from Section \ref{sec:3_3}, we assign different resolutions to frames according to their relevance. Let $K$, $M$, and $N$ be hyperparameters that define resolution thresholds based on token budgets and empirical analysis. The high, medium, and low-resolution frames are selected as follows:
\begin{equation}
\begin{split}
    \text{idx}^{\text{high}} &= \{ i \ | \ \text{rank}(i) \leq K \} \\
    \text{idx}^{\text{mid}} &= \{ i \ | K \textless \ \text{rank}(i) \leq M \} \\
    \text{idx}^{\text{low}} &= \{ i \ | M \textless \ \text{rank}(i) \leq N \}
\end{split}
\end{equation}

For each frame $\text{Frame}^{i}$, the resolution $r_i$ is assigned as follows:
\begin{equation}
    r_i = \begin{cases}
        r^{(3)} & \text{if } i \in \text{idx}^{\text{high}},\\[1mm]
        r^{(2)} & \text{if } i \in \text{idx}^{\text{mid}},\\[1mm]
        r^{(1)} & \text{if } i \in \text{idx}^{\text{low}},
    \end{cases}
\end{equation}
where $r^{(1)}$, $r^{(2)}$, and $r^{(3)}$ correspond to low, medium, and high resolutions, respectively. To transform the frame resolutions, we downsample the frames as follows: $r^{(1)} = 4r^{(2)} = 16r^{(3)}$.

This resolution adaptation has two key benefits: It ensures that critical details are preserved in high-resolution frames, and by reducing the resolution of less important frames, it minimizes the computational load, enabling the Video-LLM to process more frames within the same token budget.

\subsection{Inference with Video-LLMs}
During inference, we connect our Q-Frame to the conventional Video-LLMs. Suppose we uniformly sample $T$ candidate frames and expect $N$ frames with various resolutions as visual input for Video-LLMs. Then, the textual input and visual input are fed into respective tokenizers to extract tokens. It's important to note that frames with higher resolutions will produce more visual tokens, while frames with lower resolutions will produce fewer tokens. In the end, the textual tokens and visual tokens are connected for LLM's autoregressive generation.
%-------------------------------------------------------------------------
\section{Experiments}
In this section, we present a comprehensive evaluation of the proposed Q-Frame framework, detailing our experimental setup, main results, and ablation studies to demonstrate its effectiveness and robustness in video understanding. \textbf{Due to space limitations, additional experimental details and results are provided in the Appendix.}

\subsection{Experimental Setup}

\textbf{Baseline Models:} To validate Q-Frame, as detailed in Table~\ref{tab:baseline}, we select representative open-source models, namely VILA-V1.5 \cite{lin2024vila} and Qwen2-VL \cite{bai2023qwen}, as well as the closed-source API-based GPT-4o \cite{openai2024hello}. Q-Frame is integrated with these baselines to assess its performance relative to recent state-of-the-art Video-LLMs. In order to maintain comparability with FRAME-VOYAGE \cite{yu2024frame}, experiments using VILA-V1.5 employ QFS with fixed input frames and resolutions. Additionally, because of the closed-source nature of GPT-4o, the effect of its preprocessing on image resolution remains unknown. Consequently, the GPT-4o's experiments only utilize QFS. Our evaluation includes two experimental settings: one with a fixed number of input frames and another with a fixed token budget.

It is also noteworthy that the video understanding and multi-image understanding modes in Qwen2-VL activate distinct model weights; hence, we refer to them as Qwen2-VL-Video and Qwen2-VL, respectively.

\textbf{Benchmarks:} We evaluate Q-Frame on three widely adopted long-video benchmarks: (1) MLVU \cite{zhou2024mlvu}, a benchmark for multi-task understanding of long videos comprising 2593 tasks across nine categories (average video duration: 12 minutes); (2) LongVideoBench \cite{wu2025longvideobench}, evaluated on a validation set without subtitles, containing 1337 QA pairs (average duration: 12 minutes); and (3) Video-MME \cite{fu2024video}, which includes 2700 manually annotated QA pairs (average duration: 17 minutes).

\textbf{Implementation Details:} Following the experimental protocol in \cite{yu2024frame}, we uniformly sample 128 candidate frames and apply Q-Frame to select 8 frames (or an equivalent token budget via multi-resolution assignment) as input to the Video-LLMs. All experiments are conducted on LMMs-Eval \cite{zhang2024lmms} using 8 H100 GPUs.

\begin{table}[htbp]
    \centering
    \resizebox{0.98\linewidth}{!}{
        \begin{tabular}{lccc}
            \toprule
            \makebox[0.2\linewidth][l]{\multirow{2}{*}{\textbf{Model}}} & \makebox[0.2\linewidth][l]{\multirow{2}{*}{\textbf{Open-Source}}} & \multicolumn{2}{c}{\textbf{Q-Frame}} \\
            & & \makebox[0.2\linewidth][c]{\textbf{QFS}} & \makebox[0.2\linewidth][c]{\textbf{MRA}} \\
            \midrule
            VILA-V1.5 \cite{lin2024vila} & \checkmark & \checkmark & \texttimes \\
            GPT-4o \cite{openai2024hello} & \texttimes & \checkmark & \texttimes \\
            Qwen2-VL \cite{wang2024qwen2} & \checkmark & \checkmark & \checkmark \\
            \bottomrule
        \end{tabular}
    }
    \caption{Baseline models and experimental configurations for Q-Frame.}
    \label{tab:baseline}
\end{table}

\subsection{Main Results}

\begin{table*}[htbp]
    \centering
    \resizebox{\linewidth}{!}{
        \begin{tabular}{lcccccccc}
            \toprule
            \multirow{2}{*}{\textbf{Model}} &
            \multirow{2}{*}{\makecell{\textbf{LLM} \\ \textbf{Size}}} &
            \multirow{2}{*}{\textbf{\#Frames}} &
            \multirow{2}{*}{\textbf{MLVU}} &
            \multirow{2}{*}{\textbf{LongVideoBench}} &
            \multicolumn{4}{c}{\textbf{Video-MME$_{(wo/w\ subs)}$}} \\
            \cmidrule(lr){6-9}
            & & & & & Overall & Short & Medium & Long \\
            \multicolumn{3}{c}{\textit{Avg. Video Duration}} & \textit{12min} & \textit{12min} & \textit{17min} & \textit{1.3min} & \textit{9min} & \textit{41min} \\
            \midrule
            Video-LLaVA \cite{lin2024video}     & 7B & 8     & 47.3 & -    & 39.9 / 41.6 & 45.3 / 46.1 & 38.0 / 40.7 & 36.2 / 38.1 \\
            Qwen-VL \cite{bai2023qwen}     & 7B & 8     & -    & -    & 41.1 / 41.9 & 46.9 / 47.3 & 38.7 / 40.4 & 37.8 / 37.9 \\
            VideoChat2 \cite{li2024mvbench}     & 7B & 8     & 44.5 & -    & 39.5 / 43.8 & 48.3 / 52.8 & 37.0 / 39.4 & 33.2 / 39.2 \\
            Chat-UniVi-V1.5 \cite{jin2024chat}  & 7B & 8     & -    & -    & 40.6 / 45.9 & 45.7 / 51.2 & 40.3 / 44.6 & 35.8 / 41.8 \\
            VideoLLaMA2 \cite{cheng2024videollama} & 7B & 8   & -    & -    & 47.9 / -    & 56.0 / -    & 45.4 / -    & 42.1 / -   \\
            LLaVA-NeXT-QW2 \cite{liu2024llavanext} & 7B & 8   & -    & -    & 49.5 / -    & 58.0 / -    & 47.0 / -    & 43.4 / -   \\
            LongVILA \cite{xue2024longvila}     & 8B & 128   & -    & -    & 49.2 / -    & 60.2 / -    & 48.2 / -    & 38.8 / -   \\
            LongVA \cite{zhang2024long}         & 7B & 128   & -    & -    & 52.6 / 54.3 & 61.1 / 61.6 & 50.4 / 53.6 & 46.2 / 47.6 \\
            Video-XL \cite{shu2024video}        & 7B & 128/256 & 64.9 & -    & 55.5 / 61.0 & 64.0 / 67.4 & 53.2 / 60.7 & 49.2 / 54.9 \\
            LLaVA-OneVision \cite{li2024llava}  & 7B & *     & 64.7 & 56.3 & 58.2 / -    & - / -       & - / -       & - / -      \\
            Video-CCAM \cite{fei2024video}      & 9B & 96    & 58.5 & -    & 50.3 / 52.6 & 61.9 / 63.1 & 49.2 / 52.3 & 39.6 / 42.4 \\
            \midrule
            \rowcolor{gray!30} \multicolumn{9}{l}{\textit{Fixed input frames:}} \\  
            VILA-V1.5 \cite{lin2024vila}         & 8B & 8     & 46.3 & 47.1 & 47.5 / 50.0 & 57.8 / 61.6 & 44.3 / 46.2 & 40.3 / 42.1 \\
            \rule{0pt}{2ex} +\textsc{Frame-Voyager}\xspace \cite{yu2024frame} 
                                                 & 8B & 8     & 49.8 & -    & 50.5 / 53.6 & 60.3 / 65.0 & 47.3 / 50.3 & 43.9 / 45.3 \\
            \rule{0pt}{2ex} + \textbf{Q-Frame}   & 8B & 8     & 54.4 & 51.6 & 50.7 / 55.0 & 59.8 / 65.3 & 48.0 / 53.4 & 44.2 / 46.2 \\
             \hdashline
            GPT-4o \cite{openai2024hello}        & -  & 8     & 28.6 $\spadesuit$    & 53.3    &  61.9 / 64.5 & 69.1 / 72.1 & 62.4 / 63.8  & 54.0 / 57.9 \\
            \rule{0pt}{2ex}  + \textbf{Q-Frame} & -  &8  &29.3  &\textbf{58.6}  & \textbf{63.8} / \textbf{66.5}  & \textbf{69.9} / \textbf{73.6}  & \textbf{63.8} / \textbf{65.2}  & \textbf{57.6} / \textbf{60.8} \\
            \midrule
            \rowcolor{gray!30} \multicolumn{9}{l}{\textit{Fixed input tokens:}} \\ 
            Qwen2-VL-Video \cite{wang2024qwen2}  & 7B & 8     & 55.6    & 51.0 & 53.0 / 58.3 & 64.1 / 68.2 & 49.3 / 55.1 & 45.6 / 51.7 \\
            Qwen2-VL \cite{wang2024qwen2}   & 7B & 8  & 56.9 & 53.5 & 53.7 / 59.4 & 65.0 / 69.8 & 50.7 / 56.2 & 45.3 / 52.1 \\
            \rule{0pt}{2ex}  + \textbf{Q-Frame} & 7B & 4 + \textcolor{gray}{8} + \textcolor{gray!60}{32}  &\textbf{65.4}  &58.4 &  58.3 / 61.8 & 69.4 / 73.2 & 57.1 / 61.0 & 48.3 / 51.1 \\
            \bottomrule
        \end{tabular}
    }
    \caption{Comparing Video-LLMs with and without Q-Frame as an additional module. The experiments of LongVideoBench were performed on the validation set, excluding interleaved subtitles. Results are shown for the Video-MME benchmark under two setups: without subtitles (w/o sub.) and with subtitles (w sub.). Video-XL uses 128 frames for Video-MME and 256 frames for MLVU. LLaVA-OneVision refers to results from the official report with the well-tuned number of frames. And Video-CCAM uses 96 frames for each benchmark. \textcolor{gray}{The lighter the color, the lower the frame resolution.} The best results are shown in \textbf{bold}. $\spadesuit$: For more experimental results, refer to the appendix.}
    \label{tab:main}
\end{table*}

\subsubsection{Comparison with SOTA Methods}

Table~\ref{tab:main} presents a detailed comparison of Q-Frame applied to VILA-V1.5 \cite{lin2024vila}, Qwen2-VL \cite{wang2024qwen2}, and GPT-4o \cite{openai2024hello}, along with several recent Video-LLMs. In both fixed frame and fixed token settings, Q-Frame consistently outperforms uniform frame sampling. For example, Q-Frame significantly boosts performance in Video-MME and MLVU benchmarks, demonstrating its capability to select query-relevant frames and dynamically adjust resolutions adaptively. Compared to the baselines, Q-Frame substantially improves the SOTA performance across all three evaluated benchmarks. We highlight several key contributions:

\begin{enumerate}
    \item Our Q-Frame empowers Qwen2-VL to achieve the highest overall performance on MLVU, while GPT-4o achieved the best result in other benchmarks, confirming the effectiveness of Q-Frame.
    \item The improvements brought by Q-Frame are consistent across different Video-LLMs, whether open-source models like VILA-V1.5 and Qwen2-VL or API-based models like GPT-4o. This proves that Q-Frame is a robust, model-agnostic enhancement that can be seamlessly integrated with existing Video-LLM architectures to improve performance across diverse video understanding tasks.
    \item When Q-Frame is applied to Video-LLMs, whether open-source VILA-V1.5 and Qwen2-VL or closed-source GPT-4o, it consistently outperforms uniform frame sampling. This shows that Q-Frame is effective in frame selection and generalizes well across various Video-LLMs.

    \item Notably, VILA-V1.5 with Q-Frame selecting 8 frames outperforms FRAME-VOYAGER \cite{yu2024frame}, which employs loss-driven combinatorial optimization and trains a frame ranking model, while our Q-Frame offers a straightforward and training-free approach. 

\end{enumerate}

\subsubsection{Performance Across Video Lengths}
As shown in Table \ref{tab:lengths}, the baseline models (VILA-V1.5, GPT-4o, and Qwen2-VL) exhibit a consistent performance drop as the video duration increases, reflecting the difficulty in handling longer videos with fixed computational resources. When Q-Frame is applied, we observe significant improvements across most video lengths, particularly in longer videos, demonstrating the effectiveness of our query-aware frame selection and multi-resolution adaptation mechanisms. Here are some noteworthy insights:
\begin{enumerate}
    \item As video length increases, Video-LLMs with uniform frame sampling strategies lead to a significant drop in performance due to the inherent temporal sparsity and redundancy. Q-Frame mitigates this by intelligently selecting key frames based on their relevance to the query, yielding notable improvements, particularly for longer videos.

    % \item Balanced Resolution Scaling in Extended Videos: Q-Frame dynamically adjusts frame resolution according to their relevance to the query, thus balancing computational cost and model performance. In longer videos, where many frames are less critical, reducing the resolution of less relevant frames while maintaining higher resolutions for important frames results in a more efficient processing pipeline. For instance, GPT-4o sees a +7.3\% performance increase in the [3m, 10m] range, demonstrating the benefits of fine-tuning frame selection and resolution.

    \item Q-Frame's improvement distribution across various baselines is notable. It demonstrates a significant enhancement over VILA-V1.5, particularly in medium-length videos, while GPT-4o and Qwen2-VL show advancements aimed at long videos. This distinction may arise from the differences in the training data strategies of Video-LLMs. 

    \item The results indicate that Q-Frame significantly enhances baseline performance for most video durations. However, the performance of VILA-V1.5 in the (8s, 15] range demonstrates a slight decrease. This is because uniformly sampling 8 frames in this range results in about one frame per second, making frame selection less significant for very short videos.  

\end{enumerate}

\begin{table}[htbp]
    \centering
    \resizebox{\linewidth}{!}{
        \begin{tabular}{lccccc}
            \toprule
            \multirow{2}{*}{\textbf{Model}} &
            \multicolumn{4}{c}{LongVideoBench} \\
            \cmidrule(lr){2-5}
            \quad\ \textit{Video Duration} &
            (8s, 15s] & (15s, 1m] & (3m, 10m] & (15m, 60m] \\
            \toprule
            VILA-V1.5 \cite{lin2024vila} & 56.1 & 60.5 & 43.4 & 42.7 \\
            \rowcolor{gray!20}
            \rule{0pt}{2ex} +\textbf{Q-Frame} & $\text{52.9}^{\textcolor{blue}{-3.2}}$ & $62.8^{\textcolor{red}{+2.3}}$ & $\text{53.9}^{\textcolor{red}{+10.5}}$ & $\text{46.1}^{\textcolor{red}{+3.4}}$ \\
            \midrule
            GPT-4o \cite{openai2024hello} & 53.4 & 64.8 & 52.4 & 46.8 \\
            \rowcolor{gray!20}
            \rule{0pt}{2ex} +\textbf{Q-Frame} & $\text{60.3}^{\textcolor{red}{+6.9}}$ & $\text{68.0}^{\textcolor{red}{+3.2}}$ & $\text{59.7}^{\textcolor{red}{+7.3}}$ & $\text{54.3}^{\textcolor{red}{+7.5}}$ \\
            \midrule
            Qwen2-VL \cite{wang2024qwen2} & 53.5 & 67.6 & 51.5 & 45.9 \\
            \rowcolor{gray!20}
            \rule{0pt}{2ex} +\textbf{Q-Frame} & $\text{63.0}^{\textcolor{red}{+9.5}}$ & $\text{71.5}^{\textcolor{red}{+3.9}}$ & $\text{58.5}^{\textcolor{red}{+7.0}}$ & $\text{56.4}^{\textcolor{red}{+10.5}}$ \\
            \bottomrule
        \end{tabular}
    }
    \caption{Performance of various baselines with and without Q-Frame on LongVideoBench across different video durations.}
    \label{tab:lengths}
\end{table}

\subsubsection{Performance Across Video Tasks}
\begin{figure}[htbp]
    \centering
    \includegraphics[width=0.98\linewidth]{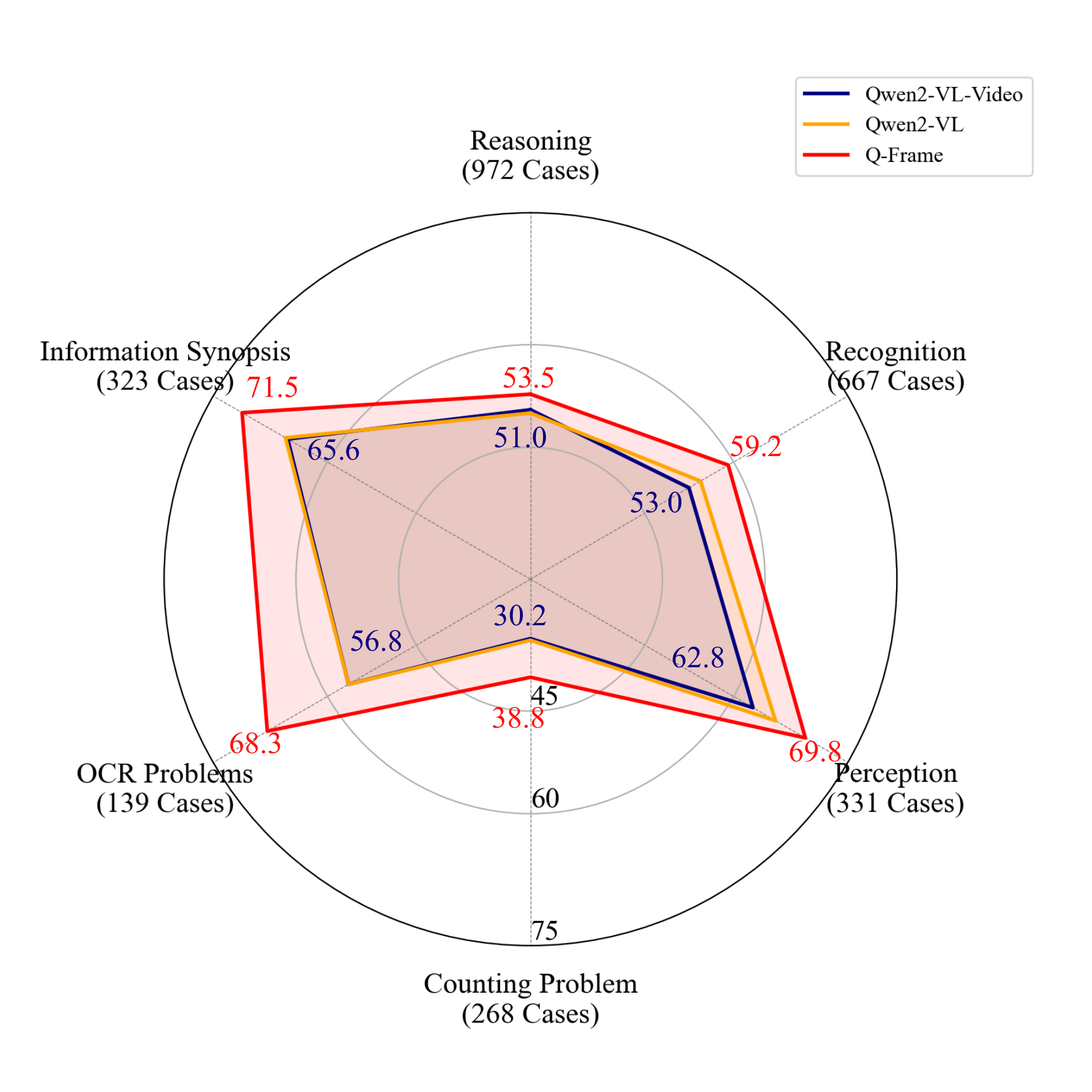}
    \vspace{-10pt}
    \caption{Accuracies (\%) of Qwen2-VL-Video, Qwen2-VL, and Q-Frame on six tasks in Video-MME. The maximum results for each task are highlighted.}
    \label{fig:radar}
\end{figure}

Figure~\ref{fig:radar} compares the accuracies of Qwen2-VL-Video, Qwen2-VL, and Q-Frame on six video tasks in Video-MME, including \emph{Reasoning}, \emph{Recognition}, \emph{Perception}, \emph{Counting Problem}, \emph{OCR Problems}, and \emph{Information Synopsis} in the Video-MME benchmark. The results reveal several important insights:

\begin{enumerate}
    \item Q-Frame enhances performance in \emph{Reasoning} tasks, where temporal context and subtle transitions are crucial. By adaptively selecting semantically relevant frames and applying multi-resolution adaptation, Q-Frame demonstrates a marked improvement over the baseline.

    \item In \emph{Recognition} tasks, Q-Frame shows significant performance improvements, suggesting that its query-aware frame selection effectively boosts the model’s capacity to capture distinctive visual spatial details.
    
    \item Q-Frame also enhances performance on \emph{Counting Problem} tasks by adaptively choosing key frames that capture all critical instances and maintain fine-grained details through multi-resolution scaling, which reduces redundancy and minimizes interference from irrelevant frames.

    \item For \emph{Perception} and \emph{OCR Problems}, the multi-resolution assignment of Q-Frame maintains fine details, which is essential for accurately identifying small objects or textual elements, resulting in superior performance.

    \item In the \emph{Information Synopsis} task, Q-Frame surpasses the baselines by effectively aggregating key scenes to create a thorough understanding of the video's narrative.
    
\end{enumerate}

Overall, these findings substantiate that Q-Frame’s adaptive strategies are particularly beneficial for tasks that demand nuanced temporal and spatial representations, thus validating the design choices underlying our framework.

\subsection{Ablation Study}
We further explore the contributions of the essential components of Q-Frame through comprehensive ablation studies on LongVideoBench using Qwen2-VL as the backbone.

\textbf{Effectiveness of Q-Frame:} Table~\ref{tab:Q-Frame} presents an ablation study of the overall Q-Frame framework, highlighting the significance of each module. The results demonstrate that the combination of QFS and MRA achieves the highest accuracy (58.4\%), confirming that the complete Q-Frame framework significantly enhances performance. Specifically, applying CLIP directly to the Top-$K$ matching frames (without QFS) yields an accuracy of 56.0\%, which is an improvement over uniform sampling. Incorporating QFS boosts performance to 57.6\%, underscoring the importance of query-aware frame selection. Additionally, using MRA alone results in a lower accuracy (52.58\%), indicating that multi-resolution adaptation alone does not fully leverage the benefits of Q-Frame. Therefore, it is evident that both QFS and MRA are essential for achieving optimal performance.

\begin{table}[htbp]
    \centering
    \resizebox{0.98\linewidth}{!}{
    \begin{tabular}{cccccc}
        \toprule
        \multicolumn{3}{c}{Sampling} & \multicolumn{2}{c}{Resolution} & \multirow{2}{*}{Acc(\%)} \\
         \makebox[0.10\linewidth][c]{Uniform} & \makebox[0.10\linewidth][c]{CLIP} & \makebox[0.10\linewidth][c]{QFS} & \makebox[0.10\linewidth][c]{Fixed} & \makebox[0.10\linewidth][c]{MRA} \\
        \bottomrule
        \checkmark & & & \checkmark & & 53.5 \\
        & \checkmark & & \checkmark & & 56.0 \\
        & & \checkmark & \checkmark & & 57.6 \\
        \checkmark & & & & \checkmark & 52.6 \\
        & \checkmark & & & \checkmark & 56.6 \\
        & & \checkmark & & \checkmark & \textbf{58.4} \\
        \bottomrule
    \end{tabular}
    }
    \caption{Ablation study of Q-Frame. CLIP denotes direct application of cross-modal matching results, and Fixed indicates a fixed resolution setting.}
    \label{tab:Q-Frame}
\end{table}

\textbf{Multi-resolution Strategy:} Table \ref{tab:MRA} highlights the effects of various resolution strategies within the MRA module. The results emphasize the advantages of multi-resolution processing, where allocating a high resolution to key frames and lower resolutions to less important frames enhances performance. The highest accuracy of 58.4\% is achieved when all three resolution categories (high, medium, and low) are utilized, illustrating that a flexible resolution strategy is ideal. When only high-resolution frames are employed, the accuracy declines to 57.6\%, and relying solely on medium or low resolutions leads to further performance degradation (55.9\% and 49.0\%, respectively). These results suggest that a balanced multi-resolution approach offers the best compromise between performance and computational efficiency.

\begin{table}[htbp]
    \centering
    \resizebox{\linewidth}{!}{
        \begin{tabular}{cccc}
            \toprule
            \multicolumn{3}{c}{Resolution} & \multirow{2}*{Acc(\%)} \\
            \makebox[0.2\linewidth][c]{High} & \makebox[0.2\linewidth][c]{Medium} & \makebox[0.2\linewidth][c]{Low} \\
            \midrule
            \checkmark & & & 57.6 \\
            & \checkmark & & 55.9 \\
            & & \checkmark & 49.0 \\
            \checkmark & \checkmark & & 57.9 \\
            \checkmark & & \checkmark & 58.1 \\
            & \checkmark & \checkmark & 56.3 \\
            \checkmark & \checkmark & \checkmark & \textbf{58.4} \\
            \bottomrule
        \end{tabular}
    }
    \caption{Ablation study of different resolution strategies within the MRA module.}
    \label{tab:MRA}
\end{table}

\textbf{Frames' resolution allocation:} Table \ref{tab:frames} analyzes how different numbers of frames with varying resolutions affect performance. To regulate the number of tokens, the quantity of frames for each resolution adheres to these guidelines:

\begin{equation}
    K + \frac{M}{4} + \frac{N}{16} = 8
\end{equation}

The optimal configuration includes 4 high-resolution frames, 8 medium-resolution frames, and 32 low-resolution frames, achieving an accuracy of 58.4\%. Further increasing the number of lower-resolution frames (up to 48) decreases accuracy to 57.4\%, highlighting the diminishing returns of adding more low-resolution frames. Moreover, due to the necessary preprocessing in MLLMs, the number of tokens has increased slightly with the increase in number, which is generally acceptable. This suggests that the Q-Frame approach benefits from balancing the number of frames of different resolutions, optimizing computational load while preserving high accuracy. 

\begin{table}[htbp]
    \centering
    \resizebox{\linewidth}{!}{
    \begin{tabular}{ccccc}
    \toprule
    \multicolumn{3}{c}{Frames} & \multirow{2}{*}{Tokens/Video} & \multirow{2}{*}{Acc(\%)} \\
     \makebox[0.12\linewidth][c]{$K$} & \makebox[0.12\linewidth][c]{$M$} & \makebox[0.12\linewidth][c]{$N$} \\
    \midrule
    8  & 0  & 0   & 2265.1  & 57.6\\
    6  & 6  & 8   & $\text{2304.1}^{\text{+1.7\%}}$  & 57.9 \\
    6  & 4  & 16  & $\text{2312.5}^{\text{+2.0\%}}$  & 58.3 \\
    4  & 8  & 32  & $\text{2344.8}^{\text{+3.5\%}}$  & \textbf{58.4} \\
    4  & 6  & 40  & $\text{2355.8}^{\text{+4.0\%}}$  & 58.2 \\
    4  & 4  & 48  & $\text{2370.1}^{\text{+4.6\%}}$  & 57.4 \\
    \bottomrule
    \end{tabular}
    }
    \caption{Ablation study of frame resolution allocation. ``Tokens/Video'' indicates the average number of tokens processed per video on LongVideoBench.}
    \label{tab:frames}
\end{table}

\subsection{Qualitative Analysis}

\begin{figure}[htbp]
    \centering
    \includegraphics[width=0.98\linewidth]{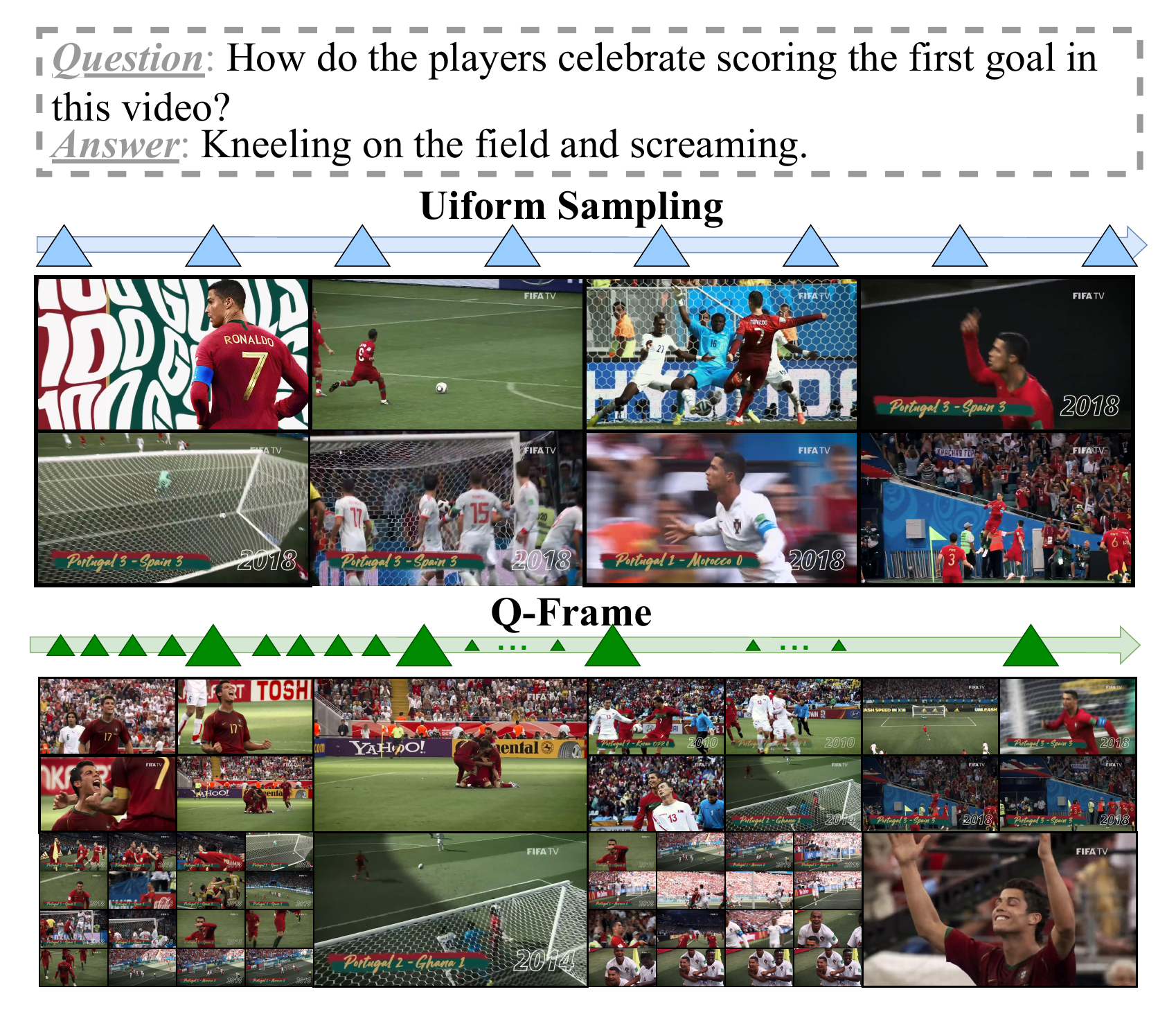}
    \caption{Case analysis from Video-MME \cite{fu2024video}. Uniform sampling captures only a limited number of frames relevant to the query. While our Q-Frame extracts more relevant frames with a variety of resolutions.}
    \label{fig:vis}
\end{figure}

Figure~\ref{fig:vis} illustrates a representative example from our qualitative analysis. In this example, the query "How do the players celebrate scoring the first goal in this video?" is answered correctly as "Kneeling on the field and screaming" when Q-Frame is applied, whereas uniform sampling fails to capture sufficient context. This comparison highlights that Q-Frame effectively selects key frames and adapts their resolutions to preserve critical temporal and spatial details, ultimately enabling more accurate and coherent responses. Additionally, with the same consumption level, our Q-Frame can process more frames and assign appropriate resolutions based on relevance.
%-------------------------------------------------------------------------
\section{Conclusion}
We introduce Q-Frame, a training-free method for efficient video understanding that dynamically selects query-aware frame sequences and adjusts frame resolutions to optimize performance while minimizing computational costs. By incorporating query-aware frame selection and multi-resolution adaptation mechanisms, Q-Frame outperforms traditional uniform sampling and keyframe selection techniques, showing superior accuracy on benchmarks such as MLVU, LongVideoBench, and Video-MME. Our experiments emphasize the essential roles of QFS and MRA in enhancing frame relevance and optimizing resolution. Q-Frame's model-agnostic, plug-and-play design offers a scalable and effective solution for long-form video understanding, paving the way for future advancements in Video-LLMs integration.
{
    \small
    \bibliographystyle{ieeenat_fullname}
    \bibliography{reference}
}

\appendix
\maketitlesupplementary

\section{Limitations}
Q-Frame enhances query-aware video understanding, but it depends on pre-trained models, lacks explicit temporal modeling, and operates within a fixed token budget. Its evaluation primarily relies on benchmarks, which highlight the need for validation in real-world applications. Future work could focus on adaptive frame budgeting, multi-modal fusion, and end-to-end optimization.

\section{Experimental Details}

\subsection{Dataset Details}

\textbf{LongVideoBench} \cite{wu2025longvideobench} is a newly introduced benchmark aimed at evaluating long-term video-language understanding for MLLMs. It comprises 3,763 web-collected videos of varying lengths (up to one hour), all featuring subtitles and encompassing a wide array of themes. This dataset is designed to assess models' capabilities to process and reason with detailed multimodal information from long-form video inputs and is notably comprehensive. It includes 6,678 human-annotated multiple-choice questions spread across 17 fine-grained categories. In this paper, we focus on the validation set without subtitles, which consists of 1,337 question-answer pairs and has an average video duration of 12 minutes.

\textbf{MLVU} \cite{zhou2024mlvu} is a new dataset designed to evaluate Long Video Understanding (LVU) performance. It addresses the limitations of existing benchmarks by providing longer video durations, a variety of video genres (including movies, surveillance footage, and cartoons), and multiple evaluation tasks. With an average video duration of 12 minutes, the benchmark includes 2,593 tasks across nine categories, delivering a thorough assessment of MLLMs' capabilities in comprehending long videos.

\textbf{Video-MME} \cite{fu2024video} is a dataset designed to enhance video understanding for Multimodal Large Language Models (MLLMs). It comprises 900 videos spanning 6 visual domains, with durations ranging from 11 seconds to 1 hour, capturing diverse contextual dynamics. All videos are manually annotated by experts, resulting in 2,700 question-answer pairs, ensuring high-quality data for model evaluation. Experiments on Video-MME will be conducted both with and without subtitles to assess the impact of multimodal inputs.

\subsection{Repeated Experiment on GPT-4o}
As shown in Table~\ref{tab:main}, the experimental results of GPT-4o on the MLVU benchmark are quite unusual. Therefore, we conducted several rounds of experiments to verify the results. The parameter configuration of the API-based model GPT-4o is detailed in Table~\ref{tab:gpt4o}. The results of the multiple rounds of experiments are presented in Table~\ref{tab:mlvu}. \underline{The underlined results are adopted in the manuscript}. This result appears to be intentional, although the specific cause remains under investigation. It might be related to how the API is invoked. The experimental conclusion indicates that, under the same experimental conditions, Q-Frame is also effective for this closed-source model.

\begin{table}[t]
    \centering
    \resizebox{\linewidth}{!}{
    \begin{tabular}{cc}
    \toprule
    \makebox[0.4\linewidth][c]{Parameter} & \makebox[0.4\linewidth][c]{Value} \\
    \midrule
    Type & Azure \\
    Model & GPT-4o \\
    Version & 2023-12-01-preview\\
    Deployment & GPT-4o \\
    \bottomrule
    \end{tabular}
    }
    \caption{Parameter configuration of the API-based model GPT-4o.}
    \label{tab:gpt4o}
\end{table}

\begin{table}[t]
    \centering
    \resizebox{\linewidth}{!}{
        \begin{tabular}{lccc}
            \toprule
            \makebox[0.2\linewidth][c]{Model} & \makebox[0.2\linewidth][c]{Round} & \makebox[0.2\linewidth][c]{\#Frames} & \makebox[0.2\linewidth][c]{Acc(\%)}  \\
            \midrule
            \multirow{3}{*}{GPT-4o \cite{openai2024hello}}        & 1  & 8     & 27.4   \\
            & 2  & 8   & \underline{28.6}    \\
            & 3 & 8  & 28.3  \\
            \midrule
            \rule{0pt}{2ex}  + \textbf{Q-Frame} & -  &8  &29.3  \\
            \bottomrule
        \end{tabular}
    }
    \caption{Comparing GPT-4o with and without Q-Frame as an additional module on MLVU benchmark. The \underline{underlined} results are adopted in the manuscript.}
    \label{tab:gpt4o}
\end{table}

\subsection{Detailed experimental results on subtasks}

\begin{table*}[h]
    \centering
    \resizebox{\linewidth}{!}{
    \begin{tabular}{lccccccccccccccccc}
    \toprule
    \multirow{2}{*}{\textbf{Model}} &\multicolumn{17}{c}{\textbf{LongVideoBench}} \\
    \cmidrule(lr){2-18}
    &TOS &S2E &E3E &S2A &SAA &O3O &T3O &T3E &O2E &T2O &S2O &TAA &T2E &E2O &SSS &T2A &SOS \\
    \midrule
    VILA-V1.5 \cite{lin2024vila} &26.0&  61.3&  52.1&  59.1&  47.2&  53.0&  44.6&  41.1&  56.3&  50.0&  45.8&  46.3&  56.9&  52.3&  23.7&  49.4&  55.6 \\ 
    \rule{0pt}{2ex} + \textbf{Q-Frame} &\textcolor{red}{31.5}  &\textcolor{red}{66.7}  &\textcolor{red}{54.3}  &\textcolor{red}{73.9}  &47.2  &\textcolor{blue}{40.9}  &\textcolor{blue}{40.5}  &\textcolor{blue}{34.2}  &\textcolor{red}{59.8}  &\textcolor{red}{56.6}  &\textcolor{red}{56.9}  &\textcolor{blue}{42.7}  &\textcolor{red}{63.1}  &\textcolor{red}{67.7}  &\textcolor{red}{24.7}  &\textcolor{red}{59.5}  &\textcolor{red}{56.8}  \\
    \midrule
    GPT-4o \cite{openai2024hello}  &34.2  &  58.2&  48.7&  67.8&  67.7&  45.2&  45.1&  62.8&  45.9&  28.9&  55.6&  63.1&  70.5&  55.6&  48.5&  50.0&  56.9  \\
    \rule{0pt}{2ex} + \textbf{Q-Frame}  &\textcolor{red}{39.7}  &\textcolor{red}{67.1}  &\textcolor{red}{67.1}  &\textcolor{red}{69.0}  &\textcolor{blue}{63.4}  &\textcolor{red}{47.9}  &\textcolor{red}{47.6}  &\textcolor{blue}{57.4}  &\textcolor{red}{54.1}  &\textcolor{red}{29.9}  &\textcolor{red}{63.0}  &\textcolor{red}{72.3}  &\textcolor{red}{76.1}  &\textcolor{red}{58.3}  &\textcolor{red}{54.5}  &\textcolor{red}{66.7}  &\textcolor{red}{66.2}
 \\
    \midrule
    Qwen2-VL-Video \cite{wang2024qwen2}  &39.7&  66.7&  59.6&  56.8&  51.4&  40.9&  45.9&  39.7&  56.3&  48.7&  51.4&  47.6&  52.3&  63.1&  35.1&  49.4&  59.3 \\
    Qwen2-VL \cite{wang2024qwen2}  &31.5&  61.3&  60.6&  53.4&  52.8&  54.5&  48.6&  41.1&  59.8&  60.5&  55.6&  47.6&  58.5&  78.5&  36.1&  51.9&  60.5  \\
    \rule{0pt}{2ex} + \textbf{Q-Frame}  &\textcolor{red}{41.1}  &\textcolor{red}{71.0}  &\textcolor{red}{66.0}  &\textcolor{red}{76.1}  &\textcolor{red}{51.4}  &\textcolor{red}{56.1}  &48.6  &41.1  &\textcolor{red}{63.2}  &\textcolor{red}{61.8}  &\textcolor{red}{65.3}  &\textcolor{red}{51.2}  &\textcolor{red}{64.6}  &\textcolor{blue}{70.8}  &\textcolor{blue}{35.1}  &\textcolor{red}{62.0}  &\textcolor{red}{66.7}  \\
    \bottomrule
    \end{tabular}
    }
    \caption{Performance of subtasks on LongVideoBench. \textcolor{red}{Red} fonts represent positive results compared to the Baseline, and \textcolor{blue}{blue} fonts represent negative results.}
    \label{tab:longvideobench}
\end{table*}

\begin{table*}[h]
    \centering
    \resizebox{\linewidth}{!}{
    \begin{tabular}{lccccccccc}
    \toprule
    \multirow{2}{*}{\textbf{Model}} &\multicolumn{9}{c}{\textbf{MLVU}} \\
    \cmidrule(lr){2-10}
    & \makebox[0.07\linewidth][c]{TR}  &\makebox[0.07\linewidth][c]{TR}  &\makebox[0.07\linewidth][c]{VS}  &\makebox[0.07\linewidth][c]{NQA}  &\makebox[0.07\linewidth][c]{ER}  &\makebox[0.07\linewidth][c]{PQA}  &\makebox[0.07\linewidth][c]{SSC}  &\makebox[0.07\linewidth][c]{AO}  &\makebox[0.07\linewidth][c]{AC} \\
    \midrule
    VILA-V1.5 \cite{lin2024vila}  &78.8  &52.0  &0.0  &56.6  &40.9  &45.8  &0.0  &29.7  &21.4  \\
    \rule{0pt}{2ex} + \textbf{Q-Frame} &\textcolor{blue}{73.1}  &\textcolor{blue}{46.5}  &0.0  &\textcolor{red}{80.3}  &\textcolor{red}{44.9}  &\textcolor{red}{56.8}  &0.0  &\textcolor{red}{30.5}  &\textcolor{red}{33.5}\\
    \midrule
    GPT-4o \cite{openai2024hello} &40.2  &24.5  &0.0  &30.7  &27.6  &26.7  &0.0  &25.5  &24.8  \\
    \rule{0pt}{2ex} + \textbf{Q-Frame}  &\textcolor{blue}{39.0}  &24.5  &0.0  &\textcolor{blue}{28.7}  &\textcolor{red}{32.1}  &\textcolor{red}{27.7}  &0.0  &\textcolor{red}{26.6}  &\textcolor{red}{25.7} 
 \\
    \midrule
     Qwen2-VL-Video \cite{wang2024qwen2} &78.4  &64.0  &0.0  &64.5 &50.6  &55.1  &0.0  &44.8  &26.2 
 \\
     Qwen2-VL \cite{wang2024qwen2} &83.3  &58.5  &0.0  &61.4  &56.8  &59.9  &0.0  &45.9  &20.4 
 \\
     \rule{0pt}{2ex} + \textbf{Q-Frame} &\textcolor{blue}{78.8}  &\textcolor{blue}{57.0}\textcolor{blue}  &0.0  &\textcolor{red}{80.3}  &\textcolor{red}{63.1}  &\textcolor{red}{69.0}  &0.0  &\textcolor{red}{53.3}  &\textcolor{red}{40.8} \\
    \bottomrule
    \end{tabular}
    }
    \caption{Performance of subtasks on MLVU. \textcolor{red}{Red} fonts represent positive results compared to the Baseline, and \textcolor{blue}{blue} fonts represent negative results.}
    \label{tab:mlvu}
\end{table*}

\begin{table*}[!h]
    \centering
    \resizebox{\linewidth}{!}{
    \begin{tabular}{lcccccc}
    \toprule
    \multirow{2}{*}{\textbf{Model}} &\multicolumn{6}{c}{\textbf{Video-MME}$_{(wo/w\ subs)}$} \\
    \cmidrule(lr){2-7}
    & \makebox[0.12\linewidth][c]{Reasoning} & \makebox[0.12\linewidth][c]{Recognition} & \makebox[0.12\linewidth][c]{Perception} & \makebox[0.08\linewidth][c]{Counting} & \makebox[0.08\linewidth][c]{OCR} & \makebox[0.15\linewidth][c]{Information Synopsis} \\
    \midrule
    VILA-V1.5 \cite{lin2024vila}  &46.5 / 49.8  &48.3 / 51.9 &58.9 / 62.6 &35.1 / 35.1   &43.9 / 51.2  &58.8 / 71.4  \\
    \rule{0pt}{2ex} + \textbf{Q-Frame} &\textcolor{red}{47.3} / \textcolor{red}{51.4} &\textcolor{blue}{46.5} / \textcolor{red}{54.5}  &\textcolor{red}{61.9} / \textcolor{red}{64.3} &\textcolor{red}{36.6} / \textcolor{red}{36.6} &\textcolor{red}{51.1} / \textcolor{red}{59.0}  &\textcolor{red}{60.1}  / \textcolor{red}{72.4}\\
    \midrule
    GPT-4o \cite{openai2024hello} &62.6 / 63.3 &60.6 / 64.0  &64.6 / 66.7 &36.6 / 46.6 &56.8 / 60.4  &81.4 / 83.3 \\
    \rule{0pt}{2ex} + \textbf{Q-Frame} & \textcolor{red}{64.9} / \textcolor{red}{66.0} & 60.6 / \textcolor{red}{65.6}  & \textcolor{red}{67.0} / \textcolor{red}{70.6}  & \textcolor{red}{38.8} / \textcolor{blue}{44.6} & \textcolor{red}{67.2} / \textcolor{red}{67.6} &\textcolor{red}{83.0} \ 83.3 \\
    \midrule
    Qwen2-VL-Video \cite{wang2024qwen2}  &51.0 / 55/6  &53.0 / 56.8  &62.8 / 64.6  &30.2 / 35.1 &56.8 / 63.3  &65.6 / 79.6 \\
    Qwen2-VL \cite{wang2024qwen2} &50.5 / 56.2  &54.9 / 57.4 &65.9 / 67.3 &30.6 / 37.3  &56.8 / 65.5 &65.9 / 79.6 \\
    \rule{0pt}{2ex} + \textbf{Q-Frame} &\textcolor{red}{53.3} / \textcolor{red}{56.5} &\textcolor{red}{59.2} / \textcolor{red}{61.3} &\textcolor{red}{69.8} / \textcolor{red}{70.4}  &\textcolor{red}{38.8} / \textcolor{red}{42.9}  &\textcolor{red}{68.3} / \textcolor{red}{67.6}  &\textcolor{red}{71.5} / \textcolor{red}{81.7} \\
    \bottomrule
    \end{tabular}
    }
    \caption{Performance of subtasks on Video-MME. \textcolor{red}{Red} fonts represent positive results compared to the Baseline, and \textcolor{blue}{blue} fonts represent negative results.}
    \label{tab:videomme}
\end{table*}

To further analyze Q-Frame's impact, we report its performance across multiple subtasks on LongVideoBench, MLVU, and Video-MME. The results indicate that Q-Frame consistently enhances model performance across most subtasks, particularly in query-dependent tasks that require adaptive frame selection.

On LongVideoBench, Q-Frame improves performance in most categories, achieving notable gains in temporal and object-centric tasks. However, some subtasks show minor drops in performance, suggesting that further optimization may be needed for specific scenarios.

For MLVU, Q-Frame significantly boosts results in non-uniform video understanding tasks such as NQA and PQA, demonstrating its effectiveness in capturing relevant frames. However, performance variations are observed in some retrieval-based tasks, likely due to the nature of the pre-trained vision-language model guiding frame selection.

In Video-MME, Q-Frame significantly improves perception, recognition, and OCR tasks, where frame relevance and resolution are critical. It also enhances reasoning and counting accuracy, though performance gains tend to be task-dependent.

Overall, these results confirm that Q-Frame effectively boosts Video-LLM performance by improving query-aware frame selection and resolution adaptation. The most significant impact is noted in tasks that require fine-grained temporal and spatial understanding.
    
\subsection{Additional Ablation Study}

\begin{table}[t]
    \centering
    \resizebox{0.75\linewidth}{!}{
    \begin{tabular}{lc}
    \toprule
    \makebox[0.2\linewidth][l]{Model} & \makebox[0.3\linewidth][c]{Acc(\%)} \\
    \midrule
      - & 53.5 \\
      CLIP \cite{radford2021learning}   & 57.8 \\
      SigLIP \cite{zhai2023sigmoid}  & 57.9 \\
      Long-CLIP \cite{zhang2024long}   & \textbf{58.4} \\
    \bottomrule
    \end{tabular}
    }
    \caption{Ablation study of the CLIP-like Model in CQR.}
    \label{tab:clip_ablation}
\end{table}

\begin{table}[!t]
    \centering
    \resizebox{0.75\linewidth}{!}{
    \begin{tabular}{cc}
    \toprule
    \makebox[0.3\linewidth][c]{$\tau$} & \makebox[0.3\linewidth][c]{Acc(\%)} \\
    \midrule
      0.5   & 57.6   \\
      0.8   & \textbf{58.4} \\
      1.0   & 58.2  \\
      1.2   & 58.0  \\
    \bottomrule
    \end{tabular}
    }
    \caption{Ablation study of the temperature parameter $\tau$ in QFS.}
    \label{tab:tau}
\end{table}

\textbf{CLIP-like Model}: CLIP's text encoder uses an absolute positional embedding restricted to 77 tokens, creating a strict limit on input token numbers \cite{radford2021learning}. Therefore, to encode longer queries, we utilize Long-CLIP \cite{zhang2024long} to overcome this limitation. Table~\ref{tab:clip_ablation} compares different CLIP-like models for Cross-modal Query Retrieval (CQR). We observe that replacing the baseline model (``-'' with no dedicated retrieval module) with standard CLIP \cite{radford2021learning} improves accuracy from 53.5\% to 57.8\%, indicating the importance of a robust vision-language alignment for retrieving semantically relevant frames. SigLIP \cite{zhai2023sigmoid} further boosts performance to 57.9\%, suggesting that more advanced training objectives can better capture cross-modal similarities. Notably, Long-CLIP \cite{zhang2024long} achieves the highest accuracy of 58.4\%, demonstrating that architectures specifically tailored for handling longer inputs and richer contexts are particularly beneficial for video retrieval. Overall, these results highlight the crucial role of a well-optimized vision-language backbone in effectively guiding frame selection for Q-Frame.

\textbf{Temperature Parameter}: Table~\ref{tab:tau} shows the effect of varying the temperature parameter $\tau$ in Query-adaptive Frame Selection (QFS). Lower values of $\tau$ (e.g., $0.5$) create a more peaked distribution, which may overlook relevant frames due to excessive exploitation, leading to a slightly lower accuracy of 57.6\%. As $\tau$ increases, the model achieves a better balance between exploration and exploitation, reaching a maximum accuracy of 58.4\% at $\tau=1.2$. These findings indicate that a moderate level of randomness introduced by the Gumbel-Max trick is crucial for identifying diverse frames; however, an excess of randomness can weaken the selection of important frames. Therefore, adjusting $\tau$ is vital for optimizing QFS performance within Q-Frame.

\textbf{Candidate Frames}: We set the number of candidate frames $T$ to 128 to ensure comparability with FRAME-VOYAGER \cite{yu2024frame}. Moreover, we conduct an ablation study to verify the impact of the number of candidate frames on Q-Frame. As shown in Table~\ref{tab:latency}, we fix the number of sampled frames to 8, and for different experimental settings of the number of candidate frames, Q-Frame can achieve stable improvement.

\textbf{Overhead}: Although Q-Frame doesn't require additional training, it introduces preprocessing overhead in the inference phase. The ablation study of the overhead of Q-Frame is shown in Table~\ref{tab:latency}. The overhead of Q-Frame focuses on calculating embeddings and performing sampling (including resolution allocation). As candidate frame embeddings can be computed in parallel, the computational cost of this part is relatively manageable. Additionally, as the number of candidate frames increases, the burden brought by MRA is almost negligible. Compared to Video-LLM's processing time, Q-Frame can introduce about 5 times more effective frames with minimal impact on time.

\begin{table}[t]
    \centering
    \resizebox{\linewidth}{!}{
        \begin{tabular}{ccccc}
        \toprule
        Candidate & Sampled &  \multicolumn{2}{c}{Latency(ms)} & \multirow{2}{*}{LongVideoBench} \\
        \cline{3-4}
        Frames & Frames & Embedding & Sampling & \\
        \midrule
        \multirow{2}{*}{64} & 8 & 76.7 & 87.1 & 56.8 \\
        & \textcolor{gray}{4} + \textcolor{gray!60}{8} + \textcolor{gray!60}{32} & 73.5 & 89.4 & 57.7 \\
        \hline
        \multirow{2}{*}{128} & 8 & 123.8 & 178.6  & 57.6 \\
        & \textcolor{gray}{4} + \textcolor{gray!60}{8} + \textcolor{gray!60}{32} & 120.1 & 182.8 & 58.4 \\
        \hline
        \multirow{2}{*}{256} & 8 & 218.0 & 361.1 & 57.8 \\
        & \textcolor{gray}{4} + \textcolor{gray!60}{8} + \textcolor{gray!60}{32} & 220.6 & 365.6 & 58.6 \\
        \bottomrule
         & 
         & 
    \end{tabular}
    }
    \caption{Ablation study of candidate frames and overhead.}
    \label{tab:latency}
\end{table}

\subsection{Additional Case Analysis}
\begin{figure*}
    \centering
    \includegraphics[width=0.85\linewidth]{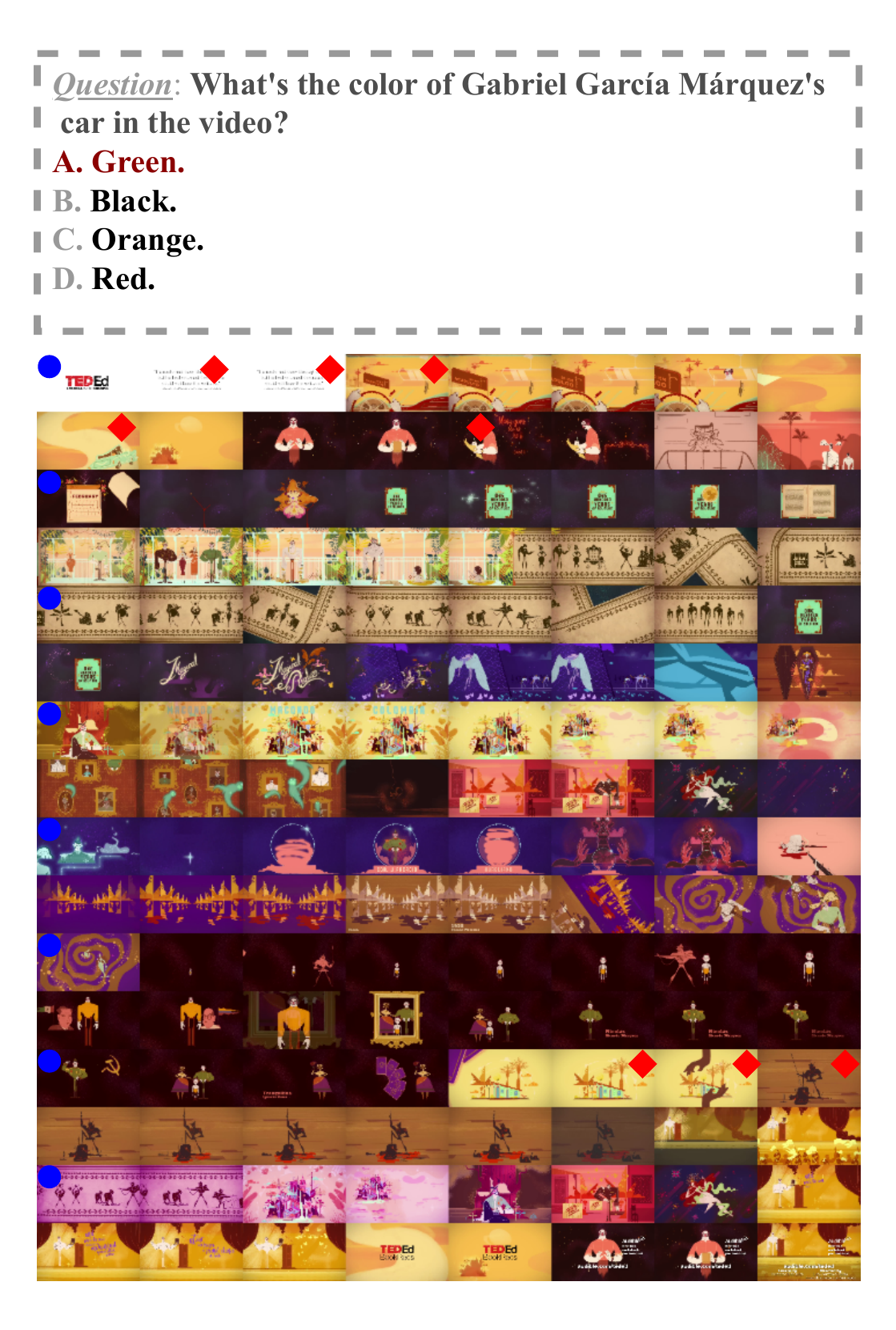}
    \caption{Case analysis from Video-MME. \includegraphics[height=0.8em]{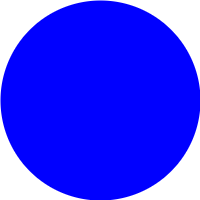}: uniform sampling;  \includegraphics[height=0.8em]{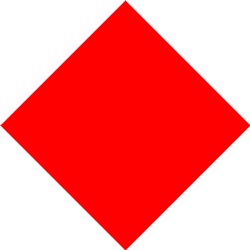}: Q-Frame. Video-LLMs with Q-Frame can capture the keyframes of the car and provide answers correctly.}
    \label{fig:case1}
\end{figure*}

\begin{figure*}
    \centering
    \includegraphics[width=0.85\linewidth]{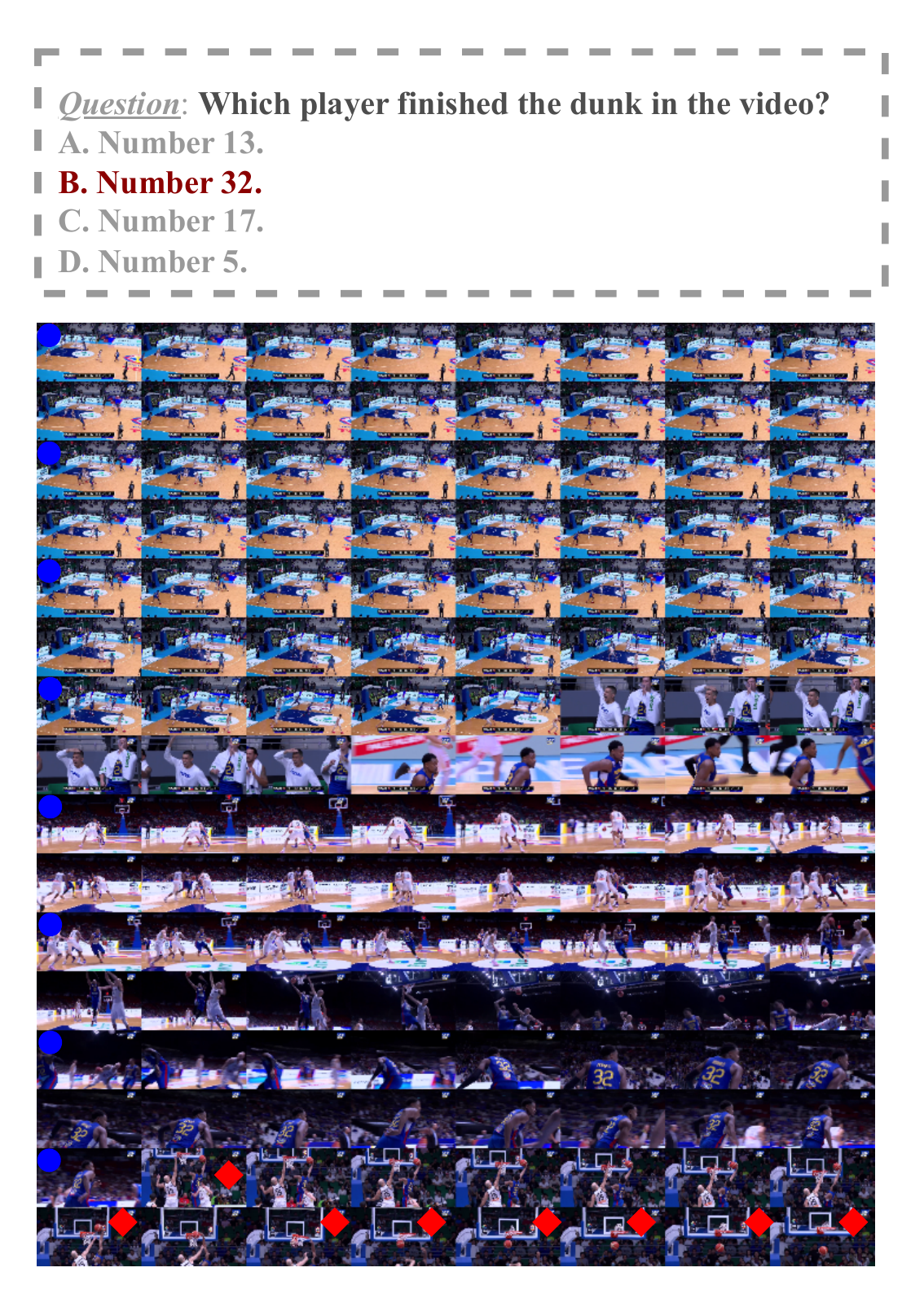}
    \caption{Case analysis from Video-MME. \includegraphics[height=0.8em]{figs/circle.png}: uniform sampling;  \includegraphics[height=0.8em]{figs/diamond.png}: Q-Frame. Video-LLMs with Q-Frame can capture the keyframes of the action of dunking and respond accordingly.}
    \label{fig:case2}
\end{figure*}

\begin{figure*}
    \centering
    \includegraphics[width=0.85\linewidth]{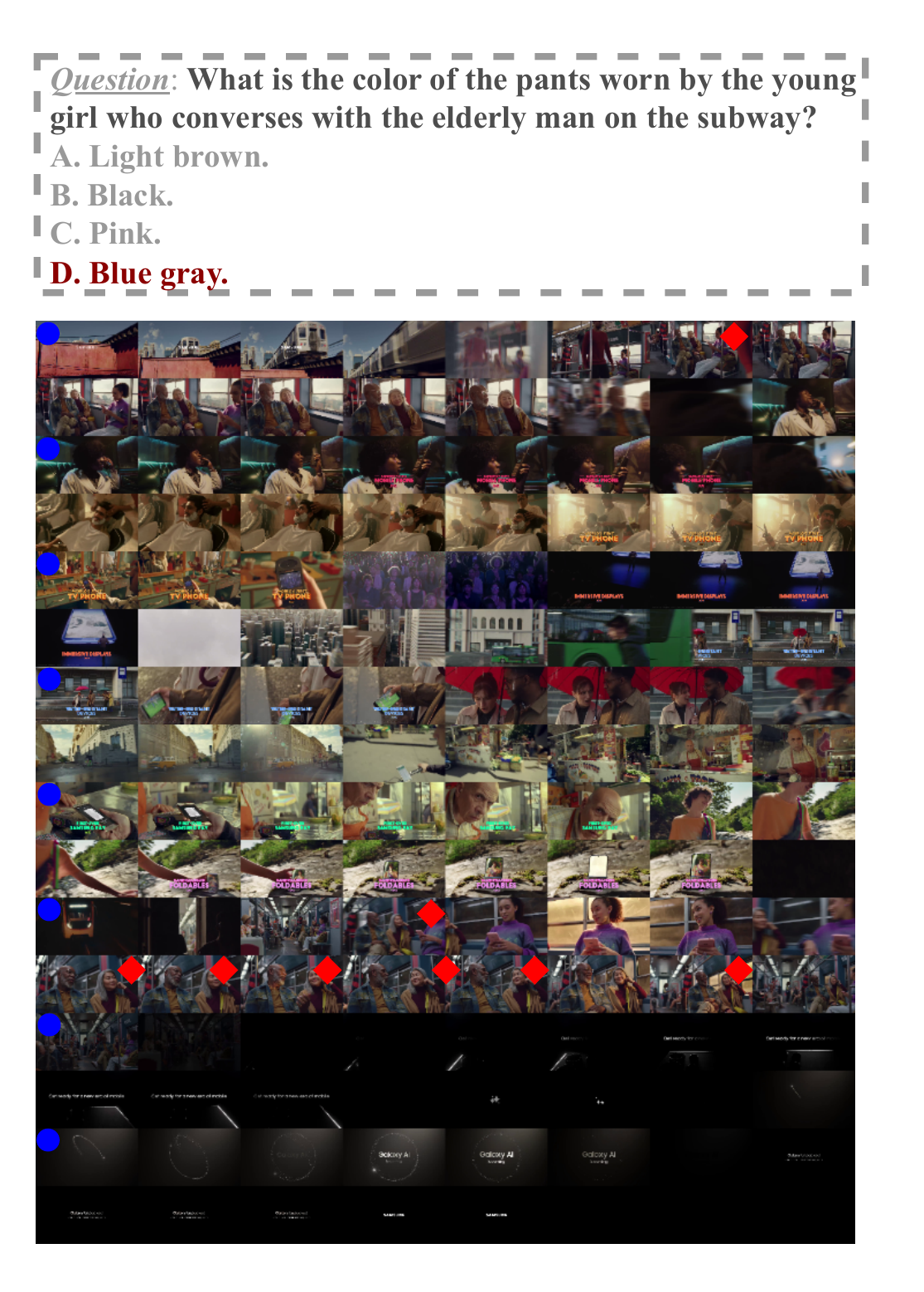}
    \caption{Case analysis from Video-MME. \includegraphics[height=0.8em]{figs/circle.png}: uniform sampling;  \includegraphics[height=0.8em]{figs/diamond.png}: Q-Frame. Video-LLMs with Q-Frame can capture the keyframes of the young girl and the elderly man and respond accordingly.}
    \label{fig:case3}
\end{figure*}

We present a comparison visualization of uniform sampling and our Q-Frame. As illustrated in Figures~\ref{fig:case1}, \ref{fig:case2}, and \ref{fig:case3}, Q-Frame more accurately identifies the video frame where the answer is located, providing a more reliable source of information for Video-LLMs. Furthermore, due to the introduction of the Gumbel-Max trick, the sampling in QFS continues over time, which solves the sparsity problem caused by sampling to some extent. 

\begin{figure*}
    \centering
    \includegraphics[width=0.85\linewidth]{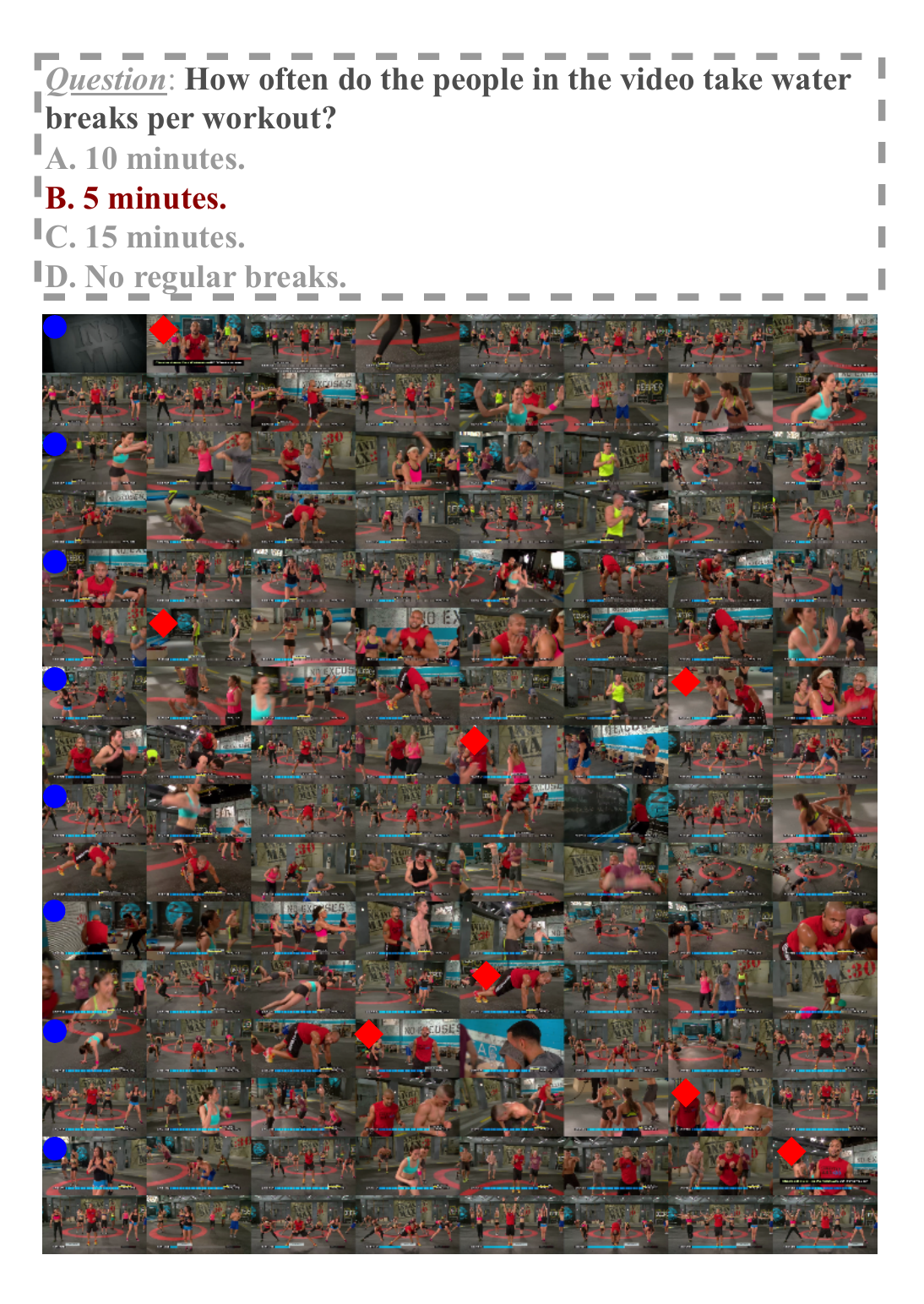}
    \caption{Case analysis from Video-MME. \includegraphics[height=0.8em]{figs/circle.png}: uniform sampling;  \includegraphics[height=0.8em]{figs/diamond.png}: Q-Frame. Neither our Q-Frame nor the uniform sampling can correctly handle such a temporal reasoning task.}
    \label{fig:case4}
\end{figure*}

\textbf{Bad Case Analysis}: However, as illustrated in Figure~\ref{fig:case4}, neither our Q-Frame nor the uniform sampling can correctly handle such a temporal reasoning task. By design, Q-Frame selects a sparse set of frames based on their semantic relevance to the query, guided by a CLIP-based matching score. This selection process doesn't preserve the sequential structure or causal relationships between events that are critical for effective temporal reasoning. Consequently, for the temporal reasoning task, even though the selected frames may be individually relevant, they often fail to capture the transitional moments or event boundaries necessary to reconstruct the full temporal logic required to answer the query accurately.

\end{document}